\documentclass[letterpaper, 10 pt, conference]{ieeeconf}
\usepackage{balance}          

\usepackage{soul} 
\usepackage{tikz}
\usetikzlibrary{calc,chains,shadows,shadows.blur,positioning}

\usepackage{svg}

\usepackage{mathtools}
\usepackage{amsmath,amsfonts}
\DeclareMathOperator*{\argmax}{arg\,max}  
\DeclareMathOperator*{\argmin}{arg\,min}  

\usepackage{algorithm}
\usepackage{algpseudocode}

\usepackage{xcolor}
\usepackage{units}
\usepackage{float}

\usepackage{subcaption}
\usepackage{hyperref}

\renewcommand{\paragraph}[1]{\noindent\textbf{#1}.}
\renewcommand{\vec}[1]{\ensuremath{\mathbf{#1}}}

\title{\LARGE \bf Minimalistic Collective Perception with Imperfect Sensors}

\author{%
  \authorblockN{Khai Yi Chin}
  \authorblockA{Dept. of Robotics Engineering\\ Worcester Polytechnic Institute, \\ Worcester, MA, USA\\ kchin@wpi.edu}
  \and
  \authorblockN{Yara Khaluf}
  \authorblockA{Dept. of Social Sciences\\ Wageningen University \& Research\\ Wageningen, NL\\ yara.khaluf@wur.nl}
  \and
  \authorblockN{Carlo Pinciroli}
  \authorblockA{Dept. of Robotics Engineering\\ Worcester Polytechnic Institute, \\ Worcester, MA, USA\\ cpinciroli@wpi.edu}
}

\begin{document}

\maketitle

\begin{abstract}
    Collective perception is a foundational problem in swarm robotics, in which the swarm must reach consensus on a coherent representation of the environment. An important variant of collective perception casts it as a best-of-$n$ decision-making process, in which the swarm must identify the most likely representation out of a set of alternatives. Past work on this variant primarily focused on characterizing how different algorithms navigate the speed-vs-accuracy tradeoff in a scenario where the swarm must decide on the most frequent environmental feature. Crucially, past work on best-of-$n$ decision-making assumes the robot sensors to be perfect (noise- and fault-less), limiting the real-world applicability of these algorithms. In this paper, we apply optimal estimation techniques and a decentralized Kalman filter to derive, from first principles, a probabilistic framework for minimalistic swarm robots equipped with flawed sensors. Then, we validate our approach in a scenario where the swarm collectively decides the frequency of a certain environmental feature. We study the speed and accuracy of the decision-making process with respect to several parameters of interest. Our approach can provide timely and accurate frequency estimates even in presence of severe sensory noise.
\end{abstract}
\section{Introduction}
\label{sec:introduction}

Constructing a coherent representation of the environment is a fundamental problem in collective robotics. From high-level representations, such as those yielded through cooperative mapping \cite{lajoie2021towards}, down to more basic representations, such as those that involve best-of-$n$ decisions, the accuracy of the result and the speed to achieve it are critical metrics for success. An important niche in this problem is that of swarms formed by \emph{minimalistic} individuals, defined by the severely limited capabilities in terms of storage, computational capabilities, and communication bandwidth. Minimalistic robots often display high levels of sensory noise, which further complicate the construction of coherent shared environment representations.

\begin{figure}
  \centering
  \includegraphics[width=.6\columnwidth]{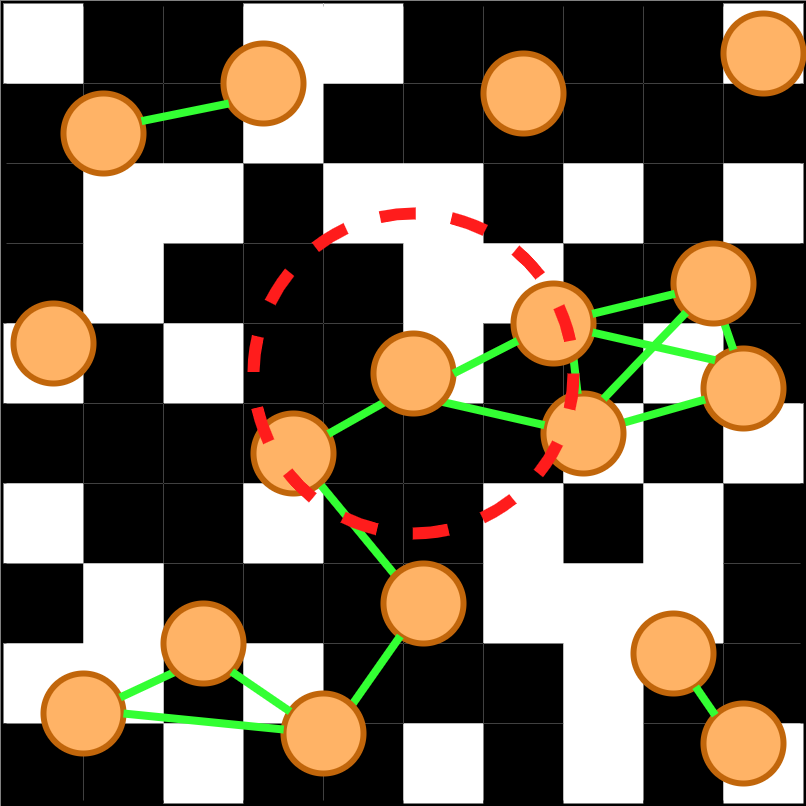}
  \caption{Environment setup for best-of-$n$ collective perception. The orange circles indicate moving robots while the green lines indicate communication links. The red dotted line around a robot denotes the communication neighborhood of that robot.}
  \label{fig:setup}
\end{figure}

In this paper, we focus on a collective perception problem cast as a best-of-$n$ decision-making process. Inspired by Valentini \textit{et al.} \cite{valentini2016perception} and Ebert \textit{et al.} \cite{ebert2020bayes}, we consider an environment with a binary feature (\textit{i.e.,} black or white floor tiles) appearing with different frequencies. Using local sensing and communication, the robots must collectively decide which feature is more frequent in the environment (see Fig. \ref{fig:setup}). Past work on this variant of collective perception assumes perfect sensing, focusing on the speed-vs-accuracy tradeoff afforded by different algorithms.

The key novelty in our problem statement is that we assume the sensors to be \emph{imperfect}, \textit{i.e.}, to provide a wrong reading with a certain non-zero probability. This could happen because of noise, faults, or as a result of ill-trained neural network classifiers. Using the theory of optimal estimation and a decentralized Kalman filter, we derive a robust and minimal collective decision-making algorithm for best-of-$n$ decision problems from first principles. Our algorithm consists of three components: \emph{(i)} The robots calculate an optimal \emph{local estimate}, using only their local sensor readings; \emph{(ii)} The robots exchange local estimates with their neighbors, and optimally aggregate them into a \emph{social estimate}; \emph{(iii)} Each robot produces an \emph{informed estimate} by combining its local and social estimates in an iterative optimization process. The informed estimates are used by the robots as the most accurate estimates of the frequency of the feature in the environment.

We study the effectiveness of our approach considering several parameters of interest: the number of robots, the density of robots in the environment, the accuracy of the sensors, and the communication topology of the robots. Results show that our collective decision-making approach offers robustness to extreme levels of sensor inaccuracy across a wide spectrum of parameter settings. Remarkably, this is produced by an algorithm with minimal requirements in terms of on-board memory ($O(1)$), computational capabilities ($O(1)$), and bandwidth ($O(1)$).


\section{Related Work}
\label{sec:related_work}

The problem of collective perception has received a wide attention in the swarm intelligence literature in both biological and artificial swarm systems \cite{ratnieks1999task,huang2003multiple}.
Best-of-$n$ decisions are commonly used to formulate collective decision-making where the options have different qualities. In Valentini \textit{et al.} \cite{valentini2016perception}, presented with an environment with black and white floor tiles, the robots must establish which tile color appear more frequently. Almansoori \textit{et al.} \cite{almansooriComparativeStudyDecision2022} performed a comparative study using a recurrent neural network-based approach against the voter model \cite{valentini2016perception}.
Reina \textit{et al.} \cite{reina2015design} proposed a framework for best-of-$n$ problems that enables engineers to move from the parameters of the emergent behavior at the macroscopic level down to the design and implementation of the individual behavior at the microscopic level. Most best-of-$n$ studies consider a binary decision problem; however, works exist that tackle more than 2 options \cite{reina2017model,bartashevich2019benchmarking}.

The most similar work to the present study is done by Ebert \textit{et al.} \cite{ebert2020bayes}, who tackled the same best-of-2 decision problem as \cite{valentini2016perception}. The authors proposed a Bayesian approach that provides probabilistic guarantees on the accuracy of the final decision, at the expense of longer times to reach a collective decision. Shan \textit{et al.} \cite{shanDiscreteCollectiveEstimation2021} extended this approach to a best-of-$n$ problem with $n > 2$. More recently, Pfister \textit{et al.} \cite{pfisterCollectiveDecisionMakingBayesian2022} extended \cite{ebert2020bayes} to consider changing floor patterns.

The works cited so far assume the information collected by the robots to always be accurate. There has been recent interest in scenarios in which a subset of robots is inaccurate, unreliable, or outright malicious in a swarm of otherwise reliable ones. Crosscombe \textit{et al.} \cite{crosscombeRobustDistributedDecisionMaking2017} proposed a 3-state voter model that allows the robots to deal with a subset of unreliable individuals with comparable computational complexity to our approach. Based on the concept of $r$-robustness, LeBlanc \textit{et al.} \cite{Leblanc2013} designed behaviors resilient to a predetermined maximum number of inaccurate individuals \cite{Guerrero2017,Saldana2017,Saulnier2017}. While $r$-robustness offers strong theoretical guarantees, these algorithms scale poorly with the size of the swarm. Strobel \textit{et al.} \cite{strobel2018managing,strobel2020blockchain} used a blockchain to counteract the presence of a significant portion of Byzantine individuals in the swarm, which share maliciously incorrect information to thwart the decision process. However, blockchain requires considerable computational resources. To our knowledge, our work is the first to consider an entire swarm of inaccurate individuals in a best-of-$n$ decision problem.
\section{Methodology}
\label{sec:methodology}

\subsection{Problem Formulation}
Our problem setup is the same as that proposed by the seminal work of Valentini \textit{et al.} \cite{valentini2016perception} and it is visually depicted in Fig. \ref{fig:setup}. A swarm of $N$ robots is scattered in a square environment. The floor is composed by tiles colored either black or white, where $f \in [0,1] \subset \mathbb{R}$ indicates the proportion of black tiles. Through an on-board ground sensor, at each time step a robot $i$ acquires a reading $z_i$, which returns either black or white. Using these readings, the robots must collectively decide the rate $f$ at which the black tiles occur in the environment.

\begin{algorithm}[t]
  \footnotesize
  \hsize=\columnwidth
  \caption{Individual robot execution}\label{alg:general}
  \begin{algorithmic}
    \Require $b \geq 0$, $w \geq 0$ \Comment{sensor accuracies}
    \Ensure $x$ \Comment{informed estimate}
    \State $t \gets 0$ \Comment{total tiles observed}
    \State $n \gets 0$ \Comment{total black tiles observed}
    \While{!Done}
    \State $\Call{Move}$
    \State $z \gets \Call{ObserveEnvironment}$
    \State $t \gets t + 1$
    \If{$z$ = Black}
      \State $n \gets n + 1$
    \EndIf
    \State $(\hat{x}, \alpha) \gets \Call{LocalEstimation}{t, n, b, w}$ \Comment{Eqs. \eqref{eq:local_sensing}, \eqref{eq:local_information}}
    \State $\mathbf{H} \gets \Call{DetectNeighbors}$ \Comment{generate list of neighbors}
    \If{$\mathbf{H}$ != empty}
      \State $\mathbf{\hat{x}_H}, \mathbf{\alpha_H} \gets \Call{Receive}{\mathbf{H}}$
      \State $\Call{Broadcast}{\mathbf{H}, \hat{x}, \alpha}$
      \State $(\bar{x}, \beta) \gets \Call{Social Estimation}{\mathbf{\hat{x}_N}, \mathbf{\alpha_N}}$ \Comment{Eqs. \eqref{eq:social_estimate}, \eqref{eq:social_information}}
    \EndIf
    \State $x \gets \Call{Informed Estimation}{\hat{x}, \alpha, \bar{x}, \beta}$ \Comment{Eq. \eqref{eq:informed_estimate}}
    \EndWhile
  \end{algorithmic}
\end{algorithm}

However, the robots have imperfect sensors, and sometimes perceive the wrong color. We define the events ``the robot observes black'' as $z_i = 1$ and ``the robot observes white'' as $z_i = 0$. Analogously, the ground truth ``the tile encountered is black'' and ``the tile encountered is white'' are captured by events $y_i = 1$ and $y_i = 0$, respectively. The sensor model is
\begin{equation}\label{eq:sensormodel}
  \begin{aligned}
    p(z_i = 1 \mid y_i = 1) &= b_i & p(z_i = 0 \mid y_i = 1) &= 1 - b_i \\
    p(z_i = 0 \mid y_i = 0) &= w_i & p(z_i = 1 \mid y_i = 0) &= 1 - w_i
  \end{aligned}
\end{equation}
where $b_i,w_i \in [0,1] \subset \mathbb{R}$ indicate the probability of a correct sensor reading for robot $i$.

The problem of collective perception can be expressed as state estimation with a consensus constraint. \textit{State estimation} is the problem of maximizing the probability of each robot's estimate of the state of the world $\hat{x}_i$ given imperfect sensor observations $\vec{z}_i$ over a time period $\Delta t$, $p(\hat{x}_i \mid \vec{z}_i; b_i, w_i)$. The consensus constraint eliminates the discrepancy between individual estimates. In its ideal form, the problem is
\begin{equation}\label{eq:cenproblem}
  \begin{aligned}
    \argmax_{\vec{\hat{x}}} \quad & \prod_i p(\hat{x}_i \mid \vec{z}_i; b_i, w_i) \\
    \text{s.t.}        \quad & \forall i \; \hat{x}_i = \bar{x}
  \end{aligned}
\end{equation}
where $\vec{\hat{x}} = [ \hat{x}_1, \dots, \hat{x}_N ]^T$ and $\bar{x}$ is the collective estimate.

\subsection{General Approach}
\label{sec:approach}

The pseudo-code of our approach is reported in Algorithm \ref{alg:general}. To make Problem \eqref{eq:cenproblem} solvable in a decentralized manner, we decompose and relax it into a two-step process. In the first step, each robot individually performs local state estimation, and in the second the robots achieve consensus:
\begin{equation}\label{eq:decproblem}
  \begin{aligned}
    \textrm{\underline{Local estimation:}} && \hat{x}_i^k &= \argmax_{\hat{x}_i} p(\hat{x}_i \mid \vec{z}_i) \\
    \textrm{\underline{Consensus:}}        && x_i^k &= \argmax_{x_i} p(x_i \mid \hat{x}_i^k, \alpha_i^k, \bar{x}_i^k, \beta_i^k)
  \end{aligned}
\end{equation}
where the exact form of the two optimization problems will be revealed in the upcoming sections. We call $x_i^k$ the \textit{informed estimate} because it combines the \textit{local estimate and confidence} $(\hat{x}_i^k, \alpha_i^k)$ (see Section \ref{sec:local_estimate}) and the \emph{social estimate and confidence} $(\bar{x}_i^k, \beta_i^k)$ (see Section \ref{sec:consensus}). Symbol $k$ indicates that Problem \eqref{eq:decproblem} is expressed as an iterative process, where the informed estimate is refined at each iteration $k$.

\subsection{Local Estimation}
\label{sec:local_estimate}

Based on the law of total probability, the probability of a robot making an observation is
\begin{align*}
    p(z = 1) &= bf + (1 - w)(1 - f) \\
    p(z = 0) &= (1 - b)f + w(1 - f),
\end{align*}
where we dropped the subscript $i$ to make the notation less cluttered. The probability of observing a total of $n \in \mathbb{Z}^+$ black tiles over a period $t \in \mathbb{Z}^+$ by a robot is characterized by the Binomial distribution
\begin{align*}
  p\bigg( \sum_{l=1}^{t} z^l = n \bigg) &= {t \choose n} \big(bf + (1 - w)(1 - f)\big)^n \cdot \\
                                        & \quad \big((1 - b)f + w(1 - f)\big)^{t - n}
\end{align*}
where $z^l$ denotes the $l$-th observation. We note that the time period $t$ can be considered as the total observations made in the discrete case. To find the estimate $\hat{x}$ that maximizes this probability, we derive the maximum likelihood estimator of $f$ through
\begin{equation*}
  \frac{\partial}{\partial f} \ln{p\bigg(\sum_{l=1}^{t} z^l = n \bigg)} = 0
\end{equation*}
which yields the solution
\begin{equation}\label{eq:local_sensing}
  f^* = \hat{x} =
  \begin{cases}
    0                                       & \text{if } n \in [0, (1-w) t], \\
    \dfrac{\dfrac{n}{t} + w - 1}{b + w - 1} & \text{if } n \in \big( (1-w)t, b t \big), \\
    1                                       & \text{if } n \in [b t, t]. \\
  \end{cases}
\end{equation}
When the sensors are perfect, \textit{i.e.,} $b=w=1$, $\hat{x}$ simplifies to $n/t$, which is the proportion of black tiles seen over $t$ observations, as intuition would suggest.

To characterize the confidence of a robot in its estimate $\hat{x}$, we use the Fisher information $\mathcal{I}(f)$, defined as
\begin{equation*}
  \alpha = \mathcal{I}(f) = -\mathbb{E} \left[ \frac{\partial^2}{\partial f^2} \ln{p\bigg( \sum_{l}^{t} z^l = n \bigg)} \,\middle\vert\, f \right].
\end{equation*}
We obtain the following result, where $q = (b + w - 1)^{2}$:
\begin{equation}\label{eq:local_information}
  \alpha =
  \begin{cases}
    \dfrac{q\big( t w^2 - 2(t - n)w + (t - n) \big)}{w^2 (w - 1)^2} & \text{if } n \in [0, (1-w) t], \\
    \dfrac{q t^3}{n(t - n)}                                         & \text{if } n \in \big( (1-w)t, b t \big), \\
    \dfrac{q(t b^2 - 2nb + n)}{b^2 (b - 1)^2}                       & \text{if } n \in [b t, t]. \\
  \end{cases}
\end{equation}

\subsection{Consensus}
\label{sec:consensus}
Once its local values $(\hat{x}_i, \alpha_i)$ has been calculated, a robot uses information from neighboring robots to generate an updated estimate, \textit{i.e.,} the \emph{informed estimate}. To this aim, a robot $i$ shares its local values with its neighbors while simultaneously receiving their local values. The informed estimate $x_i$ is solved by an optimization problem; to maximize the information in local values obtained from both robot $i$ and its neighbors, $x_i$ is defined as
\begin{equation*}
  x_i^k = \argmax_{x_i} p(x_i \mid \hat{x}_i^k, \alpha_i^k, \bar{x}_i^k, \beta_i^k)
\end{equation*}
where $(\bar{x}_i$, $\beta_i)$ are the social estimate and confidence for $i$.

In the quest to establish what the robot should communicate, we take inspiration from the theory of decentralized Kalman filtering. We assume that the underlying probability density for the local estimates is i.i.d. and Gaussian. This makes the optimization problem quadratic in $\hat{x}_i$ and $\hat{x}_j$:
\begin{align*}
  x_i &= \argmax_{x_i} \sqrt{\frac{\alpha_i}{2\pi}} \exp \bigg( -\frac{\alpha_i (x_i - \hat{x}_i)^2}{2} \bigg) \cdot \\
  & \quad \prod_{j \in \mathcal{N}_i} \sqrt{\frac{\alpha_j}{2\pi}} \exp \bigg( -\frac{\alpha_j (x_i - \hat{x}_j)^2}{2} \bigg) \\
  &= \argmin_{x_i} \frac{1}{2} \bigg( \alpha_i (x_i - \hat{x}_i)^2 + \sum_{j \in \mathcal{N}_i} \alpha_{j} (x_i - \hat{x}_{j})^2 \bigg),
\end{align*}
where $\mathcal{N}_i$ is the set of neighbors of robot $i$. Effectively, we obtain an expression akin to a decentralized Kalman filter: local estimates $\hat{x}$ from robot $i$ and its peers are fused with local confidences $\alpha$ as weights (\textit{i.e.,} the inverse of the covariances). However, differently from other approaches such as \cite{rao1993fully,bailey2007decentralised}, confidences are obtained from local estimation (in our case, Eq. \eqref{eq:local_information}). We proceed to obtain
\begin{align}\label{eq:informed_estimate}
  x_i^k &= \frac{\alpha_i^k \hat{x}_i^k + \sum_{j \in \mathcal{N}_i} \alpha_j^k \hat{x}_j^k}{\alpha_i^k + \sum_{j \in \mathcal{N}_i} \alpha_j^k}.
\end{align}
This indicates that $x_i^k$ is a weighted mean of the local estimates of the neighbors and of the robot itself. Thus, the definition of the social estimate $\bar{x}_i^k$ and the corresponding confidence $\beta_i^k$ can be taken from \eqref{eq:informed_estimate} to be
\begin{align}
  \bar{x}_i^k &= \frac{1}{\sum_{j \in \mathcal{N}_i} \alpha_j^k} \sum_{j \in \mathcal{N}_i} \alpha_j^k \hat{x}_j^k, \label{eq:social_estimate} \\
  \beta_i^k &= \sum_{j \in \mathcal{N}_i} \alpha_j^k. \label{eq:social_information}
\end{align}


Therefore, the informed estimate $x_i$ is the final estimate a robot makes of the fill ratio $f$. Since social values $(\bar{x}_i^k, \beta_i^k)$ only consider the neighbors' most recent local values $(\hat{x}_j^k, \alpha_j^k)$, bandwidth between communicating robot pairs remain fixed (\textit{O}(1)) for each time step. The on-board memory and computational requirements are also constant (\textit{O}(1)) given that each robot stores and processes only the most recent values $(n, t, \hat{x}_i^k, \alpha_i^k, \bar{x}_i^k, \beta_i^k, x_i^k)$.
\section{Experimental Evaluation}
\label{sec:experiments}

\subsection{General Setup}
\label{sec:experiments_setup}

\begin{figure*}
  \centering
  \includegraphics[width=\textwidth]{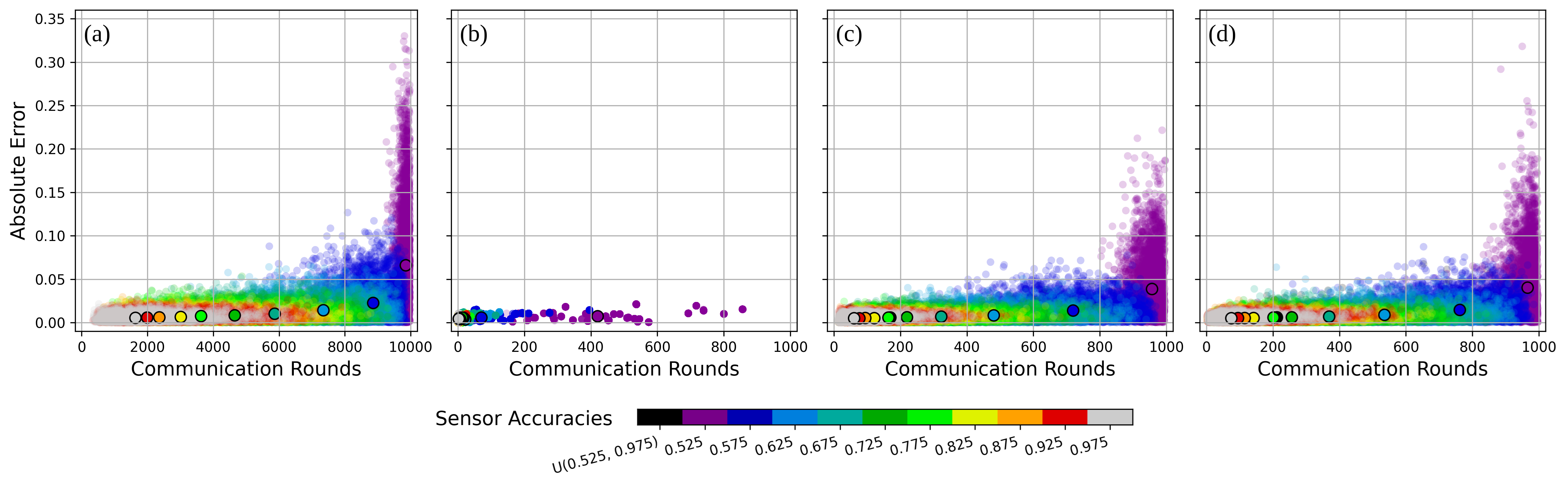}
  \caption{Estimation performance (convergence, accuracy) of 100 robots with $f = 0.55$. For each sensor accuracy, a semi-transparent dot represents a robot's estimate from one trial; there are a total of $30 \text{ trials} \cdot 100 \text{ robots} \cdot 11 \text{ sensor accuracies} = 33,000$ semi-transparent dots in each plot. Only the black dots indicate heterogeneous robots' estimates. The median performance for each sensor accuracy is denoted as a solid, outlined dot. (a) describes \emph{local} estimate performance; the remaining three describe robots' \emph{informed} estimate performance for the (b) fully-connected, (c) ring, and (d) scale-free topologies. The total amount of communication rounds shown in (a) --- where there is no communication --- indicates $T = 10,000$ observations (as if $R = 1$); this is equivalent to (b), (c), and (d) where $C = 1,000$ and $R = 10$.}
  \label{fig:static_local_informed_tfr550}
\end{figure*}

\paragraph{Communication topology}
Besides sensor accuracy, we are primarily interested in studying the role of the communication topology, a fundamental component in information and opinion spreading \cite{khaluf2018impact,rausch2020collective} which has been overlooked in previous related studies on collective perception. We consider two general scenarios: one with robots forming specific static topologies (Sec. \ref{sec:experiments_static}), and one with the robots performing random diffusion with different densities (Sec. \ref{sec:experiments_dynamic}).

\paragraph{Metrics}
We consider convergence speed, accuracy, and consensus. Because our approach aims to provide on-line estimates, we let our experiments run for $T = C \cdot R$ time steps, and establish when the robots reached convergence after a trial. We vary the number of communication rounds, $C$, and the number of observations per communication round, $R$, to control the experiment duration $T$. Unless otherwise specified, $C = 1,000$ and $R = 10$. For a single robot, we define \underline{convergence} as the point in time $K$ when $|x_i^{K} - x_i^{k}| < \delta, \forall k \geq K$, with $\delta = 0.01$. In case no convergence occurs, we record $T$ as the convergence time. As for \underline{accuracy}, we use $|x_i^K - f|$ at convergence time $K$. Our definition of \underline{consensus} is the agreement of \emph{all} the robots in their decision on the \emph{correct} target black tile fill ratio $f$. A robot `decides' by picking one out of $B$ bins that partition the entire fill ratio range, based on its informed estimate. Here we use $B = 10$, \textit{i.e.,} a best-of-10 problem; for example, a robot would select bin 8 based on its informed estimate of $0.73$, which is the correct decision for $f = 0.75$, while another robot with an informed estimate of $0.81$ would make the incorrect decision by selecting bin 9.

\paragraph{Other common parameters}
In terms of \underline{sensor accuracy}, we considered two scenarios: \emph{(i)} homogeneous accuracy with $b, w \in [0.525, 0.975], b = w$ samples with 0.05 increments; and \emph{(ii)} heterogeneous accuracy, with $b_i = w_i = U(0.525, 0.975)$, where $U(a,b)$ is the uniform distribution bounded in $[a,b]$. In terms of \underline{black tile fill ratio} $f$, we sampled the entire range $[0,1]$, but for brevity only report the cases $f \in \{0.55, 0.95\}$. Each setup $\langle\text{topology},\text{sensor accuracy},\text{fill ratio}\rangle$ was run 30 times, amounting to a total of 6,600 runs.

\subsection{Static communication topology}
\label{sec:experiments_static}

\paragraph{Topology effects}
To study the effect of specific communication topologies on collective perception, we considered four cases: fully connected, ring, line, and scale-free \cite{barabasi1999emergenceOfScaling}. In this set of experiments, once the robots are deployed to form a topology, their neighbors remain the same. At each time step, the robots acquire a sensor reading $z$ of a Bernoulli-distributed tile (parametrized by $f$). Here, robot motion is abstracted away --- each tile `encountered' is equivalent to flipping an $f$-weighted coin. Every $R = 10$ time steps (1 communication round), they also exchange messages with their immediate neighbors. We studied topologies formed by $N \in \{10, 100\}$ robots. All the static topology experiments were conducted in a custom-made Python simulator.\footnote{Available at \url{https://github.com/khaiyichin/collective_perception}.}

Fig. \ref{fig:static_local_informed_tfr550} compares the local and informed estimation performance of 100 robots in various topologies with $f=0.55$. Because the ring and line topologies yield similar results, we only discuss the ring case. We observe that performance improves with sensor accuracy across all topologies for both homogeneous and heterogeneous ($b = w = U(0.525, 0.975)$) robots. The similarity in performance between the heterogeneous robots and the $b = w = 0.75$ homogeneous robots is reasonable, since the mean of the uniform distribution is $0.75$. Robots converge onto an informed estimate much quicker than a local estimate, with improved accuracies particularly for low-quality sensors. Thus, communication improves over the individual robot estimates.

\begin{figure}[t]
  \centering
  \includegraphics[width=0.75\columnwidth]{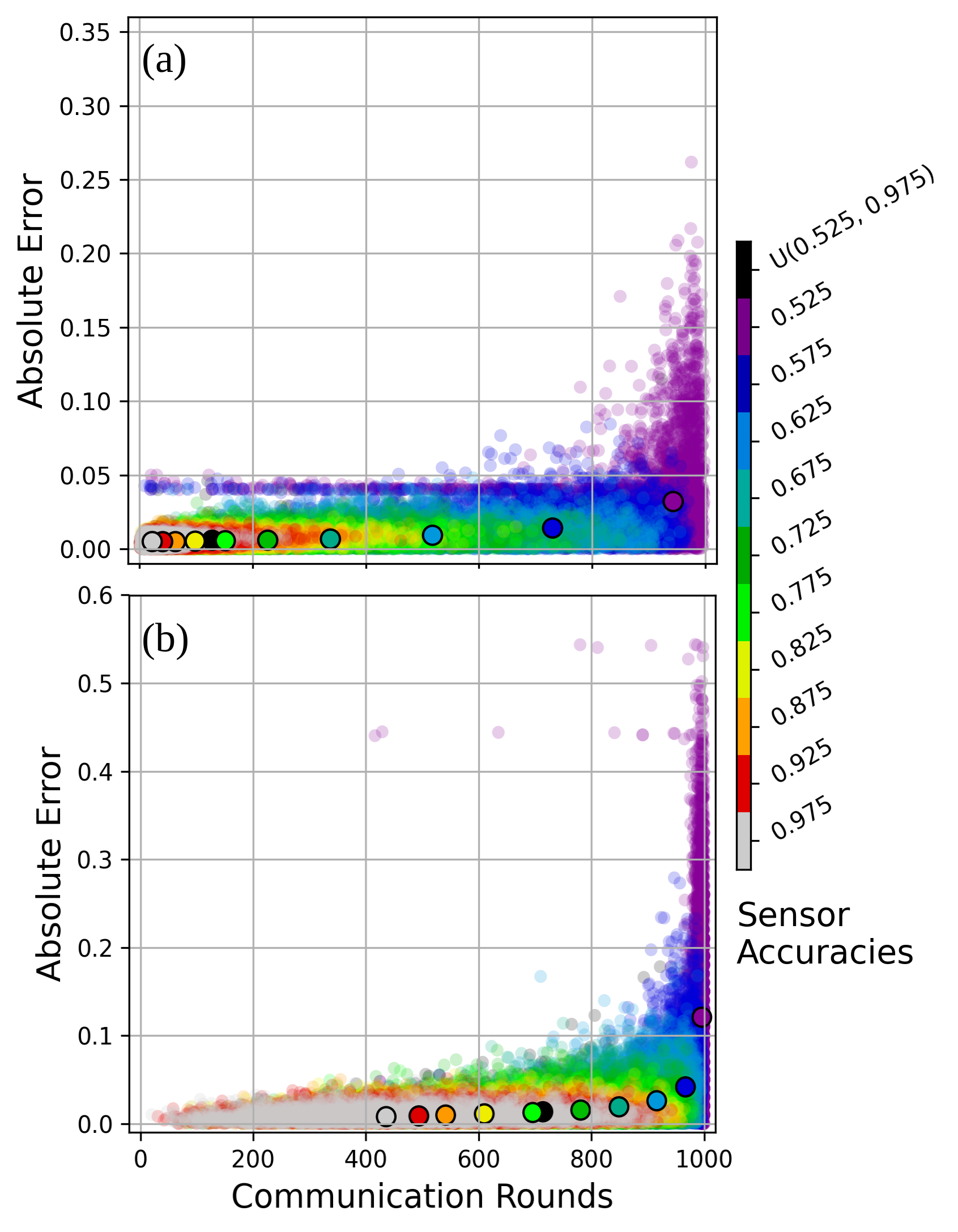}
  \caption{Informed estimation performance (convergence, accuracy) of 100 robots in the scale-free topology with: (a) $R=10$, $f = 0.95$, (b) $R = 1$, $f = 0.55$ . Note that the $y$-axis upper limit in (b) is higher than the other figures.}
  \label{fig:static_informed_combo}
\end{figure}

We also find that fully connected robots produce highly accurate estimates significantly quicker than robots in less connected networks. In fact, fully connected robots reach the same informed estimate due to how our approach computes it in Eq. \eqref{eq:informed_estimate}. With full connectivity, our approach becomes a centralized weighted average for the robots' local estimates, weighted by their respective local confidences. As for the ring and scale-free networks, topology effects (between these sparsely connected networks) on accuracy are minor: controlling for sensor accuracies and environment fill ratio ($f = 0.55$), the estimate errors remain fairly close and low, with median error values of $< 0.05$. Similar trends are observed for $f = 0.95$ in Fig. \ref{fig:static_informed_combo} (a): median estimate errors remain $< 0.05$ for all sensor accuracies and convergence speed improves with respect to sensor accuracies, while the fully-connected robots perform the best. A slight leftward shift in results when comparing $f = 0.95$ to $f = 0.55$ suggests that the former may be faster to estimate for the robots (as will be also shown in the dynamic topology results in Sec. \ref{sec:experiments_dynamic}).




\paragraph{Communication frequency effects}
In studying the effects of communication frequency, we present only the results for the scale-free network at $f=0.55$ as similar performance is observed for other networks and at $f=0.95$. Experimental results show no noticeable gain in the higher rate of information exchange ($C = 1,000$ vs. $C = 10,000$) between robots when a fixed amount of $T = C \cdot R = 10,000$ observations have been made. However, when a communication limit of $C = 1,000$ is imposed, our results indicate that premature communication slows estimate convergence. Comparing two cases with the same number of communication rounds, the robots that made $1,000$ observations ($R = 1$) in Fig. \ref{fig:static_informed_combo} (b) achieve convergence much more slowly than the robots that made $10,000$ observations ($R = 10$) in Fig. \ref{fig:static_local_informed_tfr550} (d). We thus see a benefit in prioritizing information collection before communicating estimates in our framework when communication limits arise --- either due to physical or computational hindrances.

\begin{figure*}[t]
  \centering
  \includegraphics[width=\textwidth]{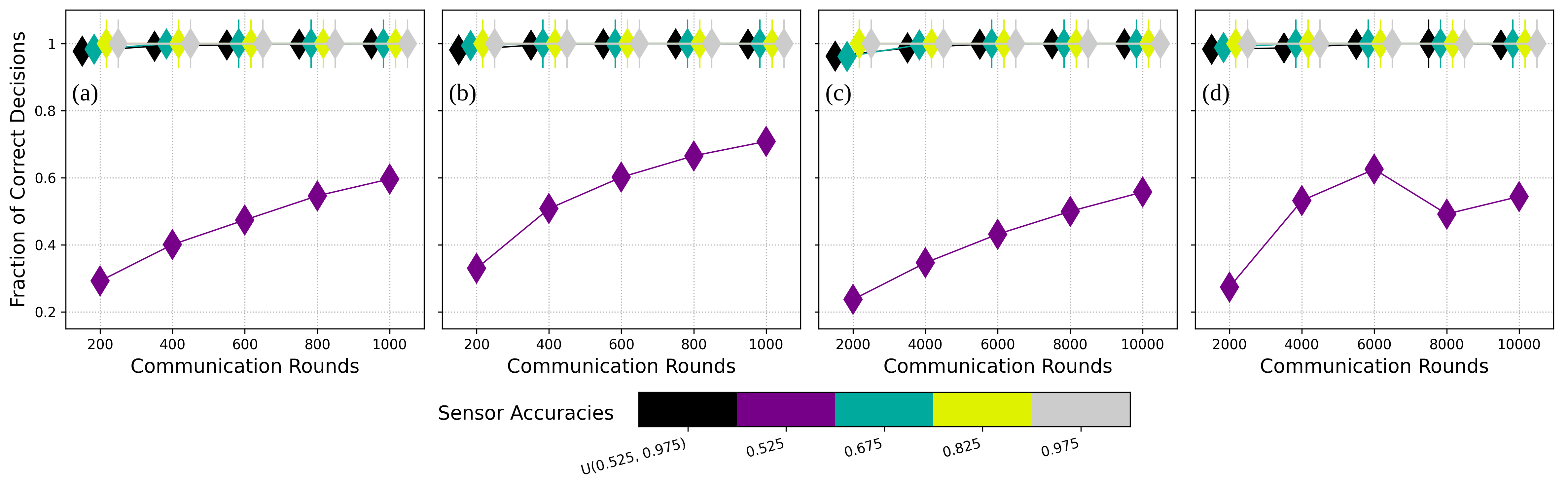}
  \caption{Swarm collective decision performance in terms of communication rounds for (a), (b) $N=100$ robots in the scale-free network and for (c), (d) $N = 25$ dynamic robots at $D = 1$. For each sensor accuracy, a diamond marker represents the fraction of robots that select the correct choice out of $B = 10$ options, based on the number of communication rounds $C$ when the decision is made. Only the black diamond markers indicate heterogeneous robots' decisions. The markers are staggered near the 100\% line to prevent obstruction of data; those that achieve 100\% correct decisions (all $30 \text{ trials} \cdot N \text{ robots} = 30N$ robots) are annotated with an additional vertical line. (a) and (c) represent $f = 0.55$ while (b) and (d) represent $f = 0.95$.}
  \label{fig:collective_decision}
\end{figure*}

\paragraph{Consensus speed}
We investigate the speed at which the swarm achieve collective consensus accurately. By looking into the fraction of robots selecting the correct bin, this tells us the duration required --- in communication rounds --- for the swarm to arrive at an accurate decision collectively. We show only results from the scale-free network as it represents the lower limits to the collective consensus behavior. Across the board, Fig. \ref{fig:collective_decision} (a) and (b) exhibit remarkable results in the swarm's collective decision-making: it eventually achieves consensus with robots of sensor accuracies $\geq 0.675$ by $1,000$ communication rounds. Assuming a reasonable duration of 1 second per communication round ($R = 10$ observations in each round), this implies that the framework provides an accurate and unanimous decision within 17 minutes, even with highly-impaired sensors. For robots in the ring and line networks, accurate collective decision is similarly achieved, often faster than the scale-free network case.

\begin{figure*}[t]
  \centering
  \includegraphics[width=\textwidth]{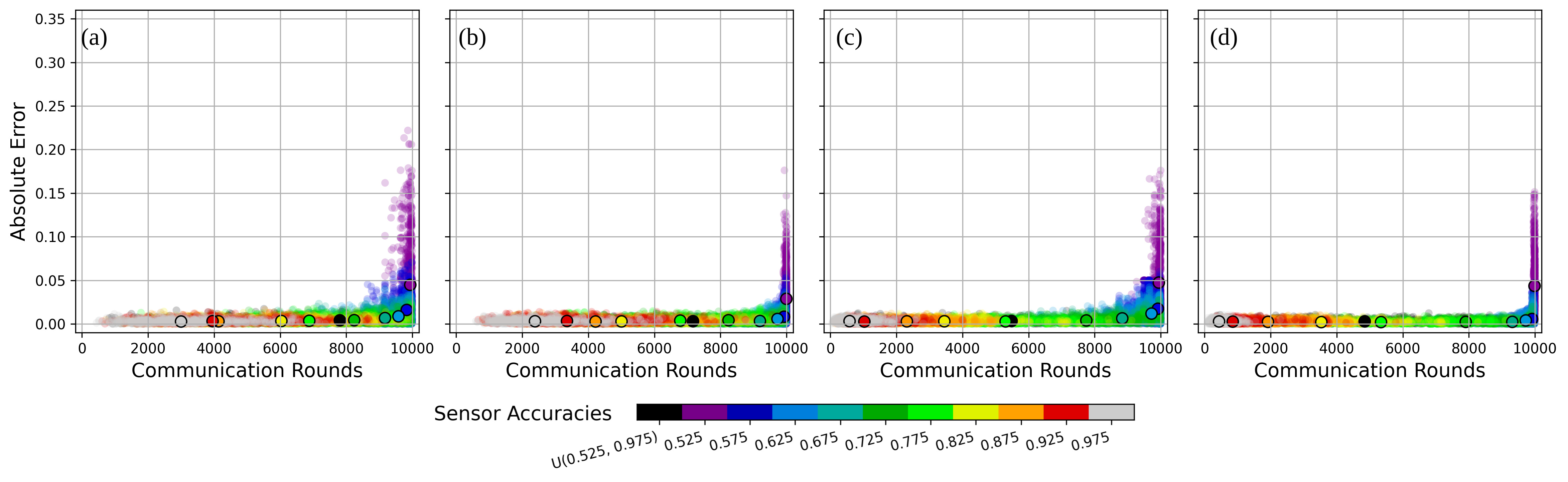}
  \caption{Informed estimation performance (convergence, accuracy) of 25 robots with varying densities and target fill ratios. Note that the number of communication rounds $C = 10,000$ is a consequence of random communication instances for the dynamic robots. (a) and (b) describe results for $D = 1$ and $D = 10$ respectively at $f = 0.55$. (c) and (d) are similar to (a) and (b) but at $f = 0.95$.}
  \label{fig:dynamic_robots}
\end{figure*}

\paragraph{Discussion}
Our results show that an increase in robot interactions leads to a quicker convergence of the estimates and quicker decision-making. However, this is true because in our approach the information exchanged is uncorrelated. This is in contrast to recent work in which robots communicate opinions \cite{crosscombeImpactNetworkConnectivity2022,talamaliWhenLessMore2021a}: a recipient robot is influenced into adopting a similar opinion as the transmitter robot's opinion, ultimately affecting the recipient's behavior. This worsened decision-making speed and often prevented consensus.

Our approach offers considerable robustness even for non-uniformly connected robots (\textit{e.g.,} scale-free topology). Our algorithm ensures \emph{accurate} collective decision-making within a reasonable amount of time ($< 17$ minutes for $R = 10$ per second) even for highly inaccurate sensors ($b = w \geq 0.675$).

\subsection{Dynamic communication topology}
\label{sec:experiments_dynamic}

\paragraph{Swarm density effects}
To study the effect of information spreading with robots in motion, we simulated swarms of $N = 25$ Khepera IV robots \cite{soares2016kheperaiv} in the ARGoS multi-robot simulator \cite{pinciroliARGoS2012}. Khepera IVs are equipped with a range and bearing communication device, ground sensors, and a ring of proximity sensors for obstacle avoidance. We set the communication range to  $r = \unit[0.7]{m}$, approximately five times the robot diameter, and let the robots move at a speed $V = \unit[0.14]{m/s}$. This is the speed that crosses a tile's diagonal length within one time step (one observation). The robots --- initialized randomly in the arena --- perform an uncorrelated random walk. The most crucial parameter in this set of the experiments is the density of the swarm in the environment. Denoting the length of the square arena side as $L$, density is defined as $D = N \pi r^2 / L^2$. For a given choice of $N$, we studied $D \in \{1, 10\}$ by changing the arena side length $L$.
For all densities $D$ and fill ratios $f$, our approach displayed similar results to those seen for static topologies'. Estimation performance increases with sensor accuracies for both homogeneous and heterogeneous robots. Density has a small influence on performance, mostly in increasing the speed of estimate convergence. As shown in Fig \ref{fig:dynamic_robots}, with a swarm density $D = 1$, sparse robot communication reduces the chances of local estimate dissemination, which explains the performance enhancement when swarm density increases to $D = 10$. Also the fill ratio $f$ has a discernible impact: both estimation accuracy and convergence speed are better for $f = 0.95$ than for $f = 0.55$. This follows the trend seen in the static topology experiments that an environment with a (close to) equally distributed feature is harder to estimate than one with a frequently present feature. Despite this, a severely impaired sensor ($b = w = 0.625$) still shows good estimation accuracy at the expense of convergence speed.


\paragraph{Consensus speed}
Using the informed estimates as decisions, we study the speed to reach collective consensus on the correct $f$. As in the static topology experiments, we let the robots pick one out of $B = 10$ bins and identify the fraction of them selecting the bin that correctly reflects $f$. Fig. \ref{fig:collective_decision} (c) and (d) show results similar to the case where robots were connected by a scale-free topology; for homogeneous swarms, the correct decision is eventually made by every robot with sensor accuracies $\geq 0.675$, while the heterogeneous swarm achieves that in a shorter time-frame when compared to some of the worst sensor accuracies $b = w \leq 0.675$. This again points to the robustness of our approach.

Because we cannot guarantee interaction between dynamic robots, communication happens whenever robots encounter one another (not unlike the static case with $R = 1$). Hence, having the robots undergo $10,000$ observations is equivalent to $C = 10,000$ communication rounds as is shown in Figs. \ref{fig:collective_decision} and \ref{fig:dynamic_robots} (it is likely that some robots have no neighbors to communicate with in a minority of those rounds). Considering a reasonable 10 Hz observation rate, \textit{i.e.,} 10 observations per second, we see that robots with sensor accuracies $\geq 0.675$ agree on the correct decision within 17 minutes, a practical duration to solve such problems.

\section{Conclusions and Future Work}
\label{sec:conclusions}
We presented an approach to collective decision-making with imperfect sensors that only require minimal computational resources. We analyzed its effectiveness by varying several parameters of interest and found that the algorithm provides resilience despite perception inaccuracies. Even highly deficient sensors ($b = w = 0.625$) can provide reasonable results at the expense of convergence speed, especially when communication is involved. Differences in communication rates affect how fast the robots make an accurate collective decision, but do not prevent the achievement of consensus. Prioritizing information collection over information exchange is a better strategy when communication instances are limited. Furthermore, having a denser swarm deployment improves the collective-decision performance; however, when communication connectivity is not guaranteed, the frequency of the environmental feature of interest may adversely affect the robots estimation.
Future work will involve several research directions. Firstly, we will study how to incorporate our framework with perception-inspired  motion instead of a diffusion process. Next, we plan to study how inhomogeneous tile distributions, e.g., clusters of black tiles, affect our approach. Finally, we intend to study the impact of robots with an incorrect estimate of their own sensor accuracy.
\section*{Acknowledgment}
This research was performed using computational resources supported by the Academic \& Research Computing group at Worcester Polytechnic Institute. This work was partially supported by grant \#W911NF2220001.


\balance
\bibliographystyle{IEEEtran}
\bibliography{references}

\begin{thebibliography}{10}
\providecommand{\url}[1]{#1}
\csname url@rmstyle\endcsname
\providecommand{\newblock}{\relax}
\providecommand{\bibinfo}[2]{#2}
\providecommand\BIBentrySTDinterwordspacing{\spaceskip=0pt\relax}
\providecommand\BIBentryALTinterwordstretchfactor{4}
\providecommand\BIBentryALTinterwordspacing{\spaceskip=\fontdimen2\font plus
\BIBentryALTinterwordstretchfactor\fontdimen3\font minus
  \fontdimen4\font\relax}
\providecommand\BIBforeignlanguage[2]{{%
\expandafter\ifx\csname l@#1\endcsname\relax
\typeout{** WARNING: IEEEtran.bst: No hyphenation pattern has been}%
\typeout{** loaded for the language `#1'. Using the pattern for}%
\typeout{** the default language instead.}%
\else
\language=\csname l@#1\endcsname
\fi
#2}}

\bibitem{lajoie2021towards}
P.-Y. Lajoie, B.~Ramtoula, F.~Wu, and G.~Beltrame, ``Towards collaborative
  simultaneous localization and mapping: a survey of the current research
  landscape,'' \emph{arXiv preprint arXiv:2108.08325}, 2021.

\bibitem{valentini2016perception}
G.~Valentini, D.~Brambilla, H.~Hamann, and M.~Dorigo, ``Collective perception
  of environmental features in a robot swarm,'' in \emph{International
  Conference on Swarm Intelligence}.\hskip 1em plus 0.5em minus 0.4em\relax
  Springer, 2016, pp. 65--76.

\bibitem{ebert2020bayes}
J.~Ebert, M.~Gauci, F.~Mallmann-Trenn, and R.~Nagpal, ``{Bayes Bots}:
  Collective bayesian decision-making in decentralized robot swarms,''
  \emph{Intl. Conference on Robotics and Automation (ICRA)}, 2020.

\bibitem{ratnieks1999task}
F.~L. Ratnieks and C.~Anderson, ``Task partitioning in insect societies. ii.
  use of queueing delay information in recruitment,'' \emph{The American
  Naturalist}, vol. 154, no.~5, pp. 536--548, 1999.

\bibitem{huang2003multiple}
M.~Huang and T.~Seeley, ``Multiple unloadings by nectar foragers in honey bees:
  a matter of information improvement or crop fullness?'' \emph{Insectes
  Sociaux}, vol.~50, no.~4, pp. 330--339, 2003.

\bibitem{almansooriComparativeStudyDecision2022}
A.~Almansoori, M.~Alkilabi, and E.~Tuci, ``A {{Comparative Study}} on
  {{Decision Making Mechanisms}} in a {{Simulated Swarm}} of {{Robots}},'' in
  \emph{2022 {{IEEE Congress}} on {{Evolutionary Computation}} ({{CEC}})}, July
  2022, pp. 1--8.

\bibitem{reina2015design}
A.~Reina, G.~Valentini, C.~Fern{\'a}ndez-Oto, M.~Dorigo, and V.~Trianni, ``A
  design pattern for decentralised decision making,'' \emph{PloS one}, vol.~10,
  no.~10, p. e0140950, 2015.

\bibitem{reina2017model}
A.~Reina, J.~A. Marshall, V.~Trianni, and T.~Bose, ``Model of the best-of-n
  nest-site selection process in honeybees,'' \emph{Physical Review E},
  vol.~95, no.~5, p. 052411, 2017.

\bibitem{bartashevich2019benchmarking}
P.~Bartashevich and S.~Mostaghim, ``Benchmarking collective perception: New
  task difficulty metrics for collective decision-making,'' in \emph{EPIA
  Conference on Artificial Intelligence}.\hskip 1em plus 0.5em minus
  0.4em\relax Springer, 2019, pp. 699--711.

\bibitem{shanDiscreteCollectiveEstimation2021}
Q.~Shan and S.~Mostaghim, ``Discrete collective estimation in swarm robotics
  with distributed {{Bayesian}} belief sharing,'' \emph{Swarm Intelligence},
  vol.~15, no.~4, pp. 377--402, Dec. 2021.

\bibitem{pfisterCollectiveDecisionMakingBayesian2022}
K.~Pfister and H.~Hamann, ``Collective {{Decision-Making}} with {{Bayesian
  Robots}} in {{Dynamic Environments}},'' in \emph{{{IROS2022}}}.\hskip 1em
  plus 0.5em minus 0.4em\relax {IEEE press}, 2022, p.~6.

\bibitem{crosscombeRobustDistributedDecisionMaking2017}
M.~Crosscombe, J.~Lawry, S.~Hauert, and M.~Homer, ``Robust {{Distributed
  Decision-Making}} in {{Robot Swarms}}: {{Exploiting}} a {{Third Truth
  State}},'' in \emph{2017 {{IEEE}}/{{RSJ International Conference}} on
  {{Intelligent Robots}} and {{Systems}}}.\hskip 1em plus 0.5em minus
  0.4em\relax {Vancouver, BC, Canada}: {IEEE}, 2017, pp. 4326--4332.

\bibitem{Leblanc2013}
H.~J. LeBlanc, H.~Zhang, X.~Koutsoukos, and S.~Sundaram, ``Resilient asymptotic
  consensus in robust networks,'' \emph{IEEE Journal on Selected Areas in
  Communications}, vol.~31, no.~4, pp. 766--781, 2013.

\bibitem{Guerrero2017}
L.~Guerrero-Bonilla, A.~Prorok, and V.~Kumar, ``Formations for resilient robot
  teams,'' \emph{IEEE Robotics and Automation Letters}, vol.~2, no.~2, pp.
  841--848, 2017.

\bibitem{Saldana2017}
\BIBentryALTinterwordspacing
D.~Saldaña, A.~Prorok, S.~Sundaram, M.~F. Campos, and V.~Kumar, ``Resilient
  consensus for time-varying networks of dynamic agents,'' in \emph{American
  {Control} {Conference} ({ACC}), 2017}.\hskip 1em plus 0.5em minus 0.4em\relax
  IEEE, 2017, pp. 252--258. [Online]. Available:
  \url{http://ieeexplore.ieee.org/abstract/document/7962962/}
\BIBentrySTDinterwordspacing

\bibitem{Saulnier2017}
\BIBentryALTinterwordspacing
K.~Saulnier, D.~Saldana, A.~Prorok, G.~J. Pappas, and V.~Kumar, ``Resilient
  {Flocking} for {Mobile} {Robot} {Teams},'' \emph{IEEE Robotics and Automation
  Letters}, vol.~2, no.~2, pp. 1039--1046, Apr. 2017. [Online]. Available:
  \url{http://ieeexplore.ieee.org/document/7822915/}
\BIBentrySTDinterwordspacing

\bibitem{strobel2018managing}
V.~Strobel, E.~Castell{\'o}~Ferrer, and M.~Dorigo, ``Managing byzantine robots
  via blockchain technology in a swarm robotics collective decision making
  scenario,'' in \emph{18th International Conference on Autonomous Agents and
  Multiagent Systems {(AAMAS 2018)}}.\hskip 1em plus 0.5em minus 0.4em\relax
  {IFAAMAS}, 2018.

\bibitem{strobel2020blockchain}
------, ``Blockchain technology secures robot swarms: A comparison of consensus
  protocols and their resilience to byzantine robots,'' \emph{Frontiers in
  Robotics and AI}, vol.~7, p.~54, 2020.

\bibitem{rao1993fully}
B.~Rao, H.~F. Durrant-Whyte, and J.~Sheen, ``A fully decentralized multi-sensor
  system for tracking and surveillance,'' \emph{The International Journal of
  Robotics Research}, vol.~12, no.~1, pp. 20--44, 1993.

\bibitem{bailey2007decentralised}
T.~Bailey and H.~Durrant-Whyte, ``Decentralised data fusion with delayed states
  for consistent inference in mobile ad hoc networks.''

\bibitem{khaluf2018impact}
Y.~Khaluf, I.~Rausch, and P.~Simoens, ``The impact of interaction models on the
  coherence of collective decision-making: a case study with simulated
  locusts,'' in \emph{International Conference on Swarm Intelligence}.\hskip
  1em plus 0.5em minus 0.4em\relax Springer, 2018, pp. 252--263.

\bibitem{rausch2020collective}
I.~Rausch, Y.~Khaluf, and P.~Simoens, ``Collective decision-making on triadic
  graphs,'' in \emph{Complex Networks XI}.\hskip 1em plus 0.5em minus
  0.4em\relax Springer, 2020, pp. 119--130.

\bibitem{barabasi1999emergenceOfScaling}
A.-L. Barabási and R.~Albert, ``Emergence of scaling in random networks,''
  \emph{Science}, vol. 286, no. 5439, pp. 509--512, 1999.

\bibitem{crosscombeImpactNetworkConnectivity2022}
M.~Crosscombe and J.~Lawry, ``The {{Impact}} of {{Network Connectivity}} on
  {{Collective Learning}},'' in \emph{Distributed {{Autonomous Robotic
  Systems}}}, F.~Matsuno, S.-i. Azuma, and M.~Yamamoto, Eds.\hskip 1em plus
  0.5em minus 0.4em\relax {Cham}: {Springer International Publishing}, 2022,
  vol.~22, pp. 82--94.

\bibitem{talamaliWhenLessMore2021a}
M.~S. Talamali, A.~Saha, J.~A.~R. Marshall, and A.~Reina, ``When less is more:
  {{Robot}} swarms adapt better to changes with constrained communication,''
  \emph{Science Robotics}, vol.~6, no.~56, p. eabf1416, July 2021.

\bibitem{soares2016kheperaiv}
J.~M. Soares, I.~Navarro, and A.~Martinoli, ``The khepera {{IV}} mobile robot:
  {{Performance}} evaluation, sensory data and software toolbox,'' in
  \emph{Robot 2015: {{Second}} Iberian Robotics Conference}, L.~P. Reis, A.~P.
  Moreira, P.~U. Lima, L.~Montano, and V.~{Mu{\~n}oz-Martinez}, Eds.\hskip 1em
  plus 0.5em minus 0.4em\relax {Cham}: {Springer International Publishing},
  2016, pp. 767--781.

\bibitem{pinciroliARGoS2012}
C.~Pinciroli, V.~Trianni, R.~O'Grady, G.~Pini, A.~Brutschy, M.~Brambilla,
  N.~Mathews, E.~Ferrante, G.~Caro, F.~Ducatelle, M.~Birattari, L.~Gambardella,
  and M.~Dorigo, ``\BIBforeignlanguage{Undefined/Unknown}{Argos: a modular,
  parallel, multi-engine simulator for multi-robot systems},''
  \emph{\BIBforeignlanguage{Undefined/Unknown}{Swarm Intelligence}}, vol.~6,
  no.~4, pp. 271--295, 2012, impact Factor: 3.12.

\end{thebibliography}

\end{document}